\documentclass{article}

\usepackage{spconf,amsmath,graphicx,mathabx}

\title{Coarse-to-fine volumetric segmentation of teeth in Cone-Beam CT}

\name{Matvey Ezhov, Adel Zakirov, Maxim Gusarev}
  
\address{Diagnocat \\
Moscow, Russia \\
matvey, adel, m.gusarev@diagnocat.com}

\name{
    Matvey Ezhov$^{\convolution}$ \qquad 
    Adel Zakirov$^{\star}$ \qquad 
    Maxim Gusarev$^{\varstar}$
}

\address{
    Diagnocat \\ Moscow, Russia \\
    $^{\convolution}$ 
    matvey@diagnocat.com\qquad 
    $^{\star}$ 
    az@diagnocat.com\qquad 
    $^{\varstar}$
    m.gusarev@diagnocat.com
}
  
\begin{document}

\maketitle

\begin{abstract}
We consider the problem of localizing and segmenting individual teeth inside 3D Cone-Beam Computed Tomography (CBCT) images. To handle large image sizes we approach this task with a coarse-to-fine framework, where the whole volume is first analyzed as a $33$-class semantic segmentation (adults have up to $32$ teeth) in coarse resolution, followed by binary semantic segmentation of the cropped region of interest in original resolution. To improve the performance of the challenging 33-class segmentation, we first train the Coarse step model on a large weakly labeled dataset, then fine-tune it on a smaller precisely labeled dataset. The Fine step model is trained with precise labels only. Experiments using our in-house dataset show significant improvement for both weakly-supervised pretraining and for the addition of the Fine step. Empirically, this framework yields precise teeth masks with low localization errors sufficient for many real-world applications.
\end{abstract}

\begin{keywords}
CBCT, segmentation, volumetric data, 3D CNN, FCN, weak supervision, fine tuning
\end{keywords}

\section{Introduction}
\label{sec:intro}
In recent years, machine learning has been successfully applied to various medical imaging problems, but the use of it within the field of dental radiography remains limited, especially with 3D CBCT scans. In this paper, we present methods for 3D tooth segmentation. We start by training a model to predict coarse (downscaled) segmentation using a large weakly labeled dataset (it should be noted that it is currently impossible to process typical CBCT scans in original resolution with sufficiently large networks). Then, we fine-tune this model on a smaller, precisely labeled dataset while still predicting coarse masks. Finally, we train a separate model to predict high-resolution segmentation inside cropped regions of interest (RoI) based on coarse masks of individual teeth, using only the precisely labeled dataset.

The main contribution of our work is two-fold. First, we show how using a combination of a large dataset with weak labels and a small dataset with precise labels can yield substantial improvement in performance compared to the use of a small precisely labeled dataset only. Second, we show that it is possible to train effective segmentation pipelines that are able to localize and precisely segment at least $32$ distinct anatomical structures.

\section{Related work}
\label{sec:relatedwork}
Machine learning is rapidly transforming the medical imaging application landscape \cite{dlsurvey}. Convolutional neural network architectures such as U-Net \cite{unet} for 2D and V-Net \cite{vnet} and 3D U-Net \cite{3dunet} for 3D images have proven effective for anatomic structure segmentation on medical images, as well as \cite{p1,p2,p3}. Generative Adversarial Networks \cite{gans} are applied for medical image synthesis \cite{ganssynth} and domain adaptation \cite{gansda1, gansda2}. In \cite{pyramid} authors propose an approach that utilizes auto-context to perform semantic segmentation at high resolutions in a multi-scale pyramid of stacked 3D FCNs. A network architecture called the label refinement network is shown in \cite{Islam2017LabelRN} to predict segmentation labels in a coarse-to-fine fashion at several resolutions.

To our knowledge, applications of deep learning to dental imaging have been relatively sparse. \cite{bench} evaluates different deep learning methods for dental X-ray analysis. In \cite{dl1, dl2} authors used deep convolutional neural networks for tooth type classification and labeling on dental CBCT images. \cite{dl3,dl4} apply deep neural networks to caries detection. The U-Net architecture was trained to segment dental X-ray images in the Grand Challenge for Computer-Automated Detection of Caries in Bitewing and won by a large margin. \cite{dl5} presents an overview of the machine and deep learning techniques in dental image analysis.

\section{Datasets}
\label{sec:dataset}
We used a dataset of depersonalized 3D CBCT head scans with isotropic voxel spacing from $0.15$ mm to $0.4$ mm and fields of view varying from just a few teeth to capturing the whole head. Typical image size is $400\times600\times600$. Voxel values represent approximate Hounsfield Units, which measure radiodensity inside a voxel volume (in CBCTs it is not guaranteed that intensity values fully correspond to the Hounsfield scale). Care was taken to include scans produced by devices manufactured by $6$ different companies. We believe this dataset is representative of the majority of dental CBCT images obtained "in the wild". For those scans, two sets of labels were collected: a sparse axial bounding box annotation set and a 3D voxel-wise mask set.

A set of $815$ studies was annotated by $4$ specialists using axial bounding boxes. The process consisted of the specialist selecting $3$ to $7$ axial slices per upper and lower jaw, drawing a bounding box around the tooth axial profile, and entering a tooth number. Although bounding boxes are sufficient for learning tooth detection on 2D axial slices, we consider them weak labels for 3D segmentation in the sense that they provide hints to location and size of the tooth, but not its precise boundary. Approximate time to create such annotation is 30 minutes per CBCT volume.

Then, a set of $120$ studies was annotated by $5$ specialists using per-voxel label assignment. Specialists used MITK software \cite{mitk} that allows quick delineation of tooth view in three planes and exploits subsequent 3D contour interpolation. Then two different specialists validated segmentation results using ITK-SNAP software \cite{itksnap} that allows detailed 2D and 3D inspection of the previously obtained mask, as well as its own manual and semi-automatic segmentation tools. We found MITK to be more labor-efficient, while ITK-SNAP is easier to use, more performant, and memory-efficient. Approximate time to annotate a single volume this way is $6$ hours.

Voxel-wise segmentation annotation requires a lot of time, attention, and proficiency with 3D software from a specialist, making it difficult and expensive to build a large dataset. However, it can yield very precise masks. On the other hand, bounding box annotation on axial slices is a very fast and simple procedure. However, to get volumetric masks from sparse axial bounding boxes we had to use a set of heuristic procedures, which led to artifacts and incorrect labels on resulting masks. It takes approximately $30$ minutes for a specialist to segment one 3D CBCT scan with bounding boxes, while voxel-wise segmentation takes about 6 hours and requires powerful hardware and software training.

\begin{figure}[htb]

\begin{minipage}[b]{1.0\linewidth}
  \centering
  \centerline{\includegraphics[width=8.5cm]{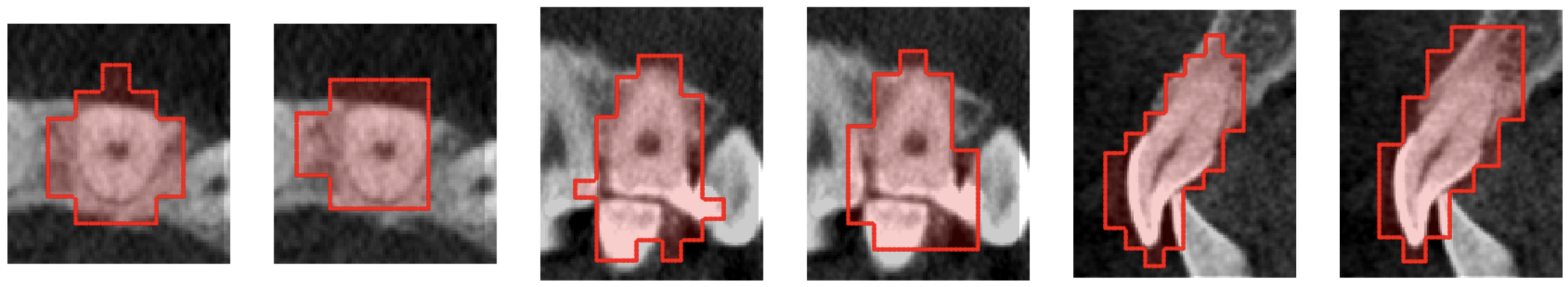}}
  {(a) Weak manual segmentation and prediction of the Coarse model trained on that data.}\medskip
\end{minipage}
\begin{minipage}[b]{1.0\linewidth}
  \centering
  \centerline{\includegraphics[width=8.5cm]{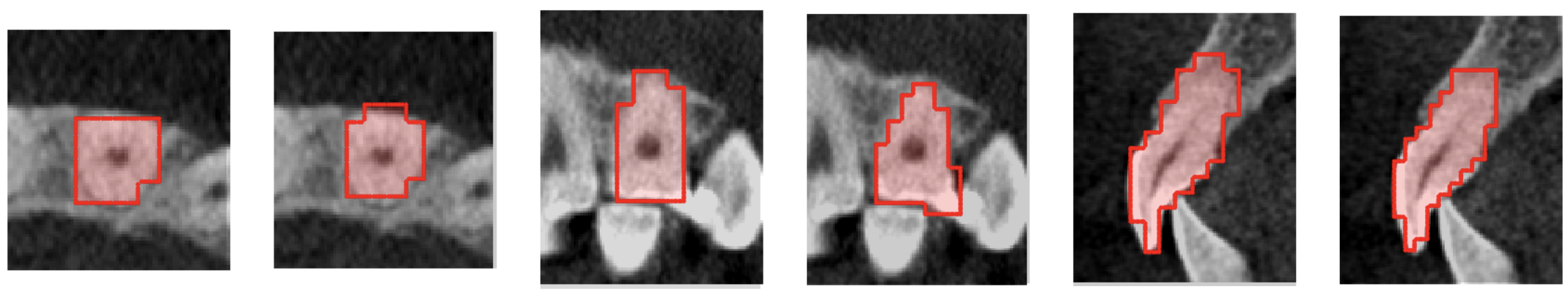}}
  {(b) Downsampled precisely labeled masks and fine-tuned the Coarse model predictions.}\medskip
\end{minipage}
\hfill
\begin{minipage}[b]{1.0\linewidth}
  \centering
  \centerline{\includegraphics[width=8.5cm]{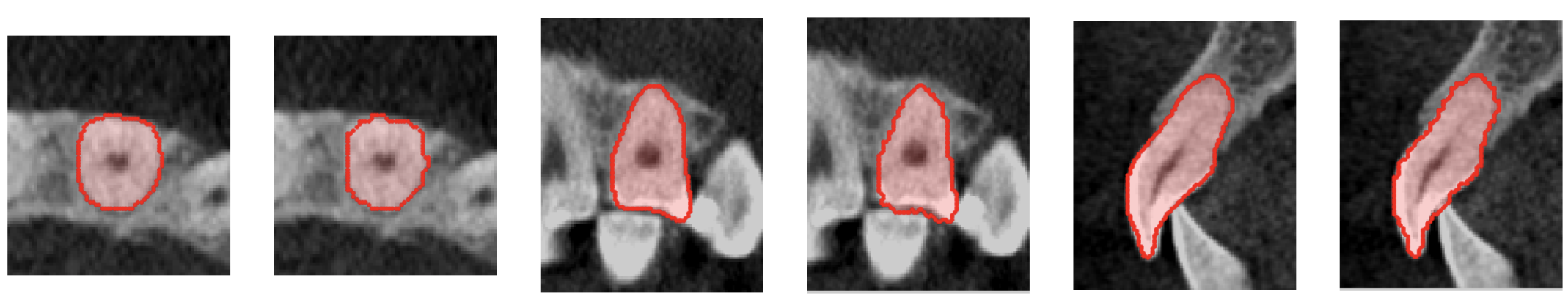}}
  {(c) Original precise labeled masks and results of the Fine model.}\medskip
\end{minipage}
\caption{Examples of the model trained on weakly labeled data, the Coarse model and Fine model on 2D axial, sagittal and frontal slices. Ground truth labels are on the right and corresponding model predictions are on the left. We propose a pipeline for training precise 3D tooth segmentation model: (a) predicting coarse (downscaled) segmentation using large weakly labeled dataset, (b) fine-tuning this model on a smaller, precisely labeled downscaled dataset while still predicting coarse masks, (c) high-resolution segmentation inside cropped RoI's based on coarse masks of individual teeth, using small precisely labeled dataset only. }
\label{fig:res}
\end{figure}

\begin{table*}[t]
    \caption{\textbf{Results.} Here, we present the evaluation of the Coarse-only model and the Coarse-to-Fine pipeline on different types and amounts of data. \textbf{(1)} evaluates the Coarse-only model on full weakly labeled dataset ($760$ cases); \textbf{(2)} evaluates the Coarse-only model on full precisely labeled dataset (93 cases); \textbf{(3)} evaluates the Coarse-only model pretrained on full weakly labeled dataset, then fine-tuned on increasing amounts of precisely labeled data; \textbf{(4)} evaluates a regime similar to (3) with the addition of the Fine model (separate model working in original resolution), where the Fine model is trained on the same part of precisely labeled data as the Coarse model is. The time cost is an estimated amount of annotator-hours required to obtain a dataset of this size (excluding development and test sets). All evaluations were performed on the hold-out test set of $20$ precisely labeled cases.}
  \label{results}
  \centering
  \setlength{\tabcolsep}{5pt}
  \begin{tabular*}{\textwidth}{ @{\extracolsep{\fill}} cccccccc} 
    \hline
    & Model & Weak dataset \% & Precise dataset \% & Precise dataset \% & Time cost, & ASD & IoU \\
    & & used for Coarse step & used for Coarse step & used for Fine step & hours & & \\
    \hline
    (1) & Coarse-only & 100 & 0 & 0 & 380 & 0.66 & 0.61 \\
    (2) & Coarse-only & 0 & 100 & 0 & 558 & 1.72 & 0.58 \\
    (3.1) & Coarse-only & 100 & 5 & 0 & 410 & 0.43 & 0.67 \\
    (3.2) & Coarse-only & 100 & 10 & 0 & 440 & 0.36 & 0.70 \\
    (3.3) & Coarse-only & 100 & 25 & 0 & 524 & 0.33 & 0.72 \\
    (3.4) & Coarse-only & 100 & 50 & 0 & 662 & 0.28 & 0.75 \\
    (3.5) & Coarse-only & 100 & 100 & 0 & 938 & 0.24 & 0.78 \\
    (4.1) & Coarse-to-Fine & 100 & 5 & 5 & 410 & 0.28 & 0.91 \\
    (4.2) & Coarse-to-Fine & 100 & 10 & 10 & 440 & 0.25 & 0.92 \\
    (4.3) & Coarse-to-Fine & 100 & 25 & 25 & 524 & 0.20 & 0.93 \\
    (4.4) & Coarse-to-Fine & 100 & 50 & 50 & 662 & 0.19 & 0.93 \\
    (4.5) & Coarse-to-Fine & 100 & 100 & 100 & 938 & \textbf{0.17} & \textbf{0.94} \\
    \hline
  \end{tabular*}
\end{table*}

\section{Methods}
\label{sec:methods}
Our approach consists of the following steps: (1) Preprocessing incoming volumetric image; (2) Coarse step weakly-supervised pretraining; (3) Coarse step fine-tuning on downscaled precise masks; (4) Fine step training in original resolution on original precise masks (Figure 1).

For preprocessing we tried different methods, including removing outlier intensities by clipping to high and low percentiles, normalizing inside $0$ to $1$ range, histogram equalization, applying a fixed intensity window width and window level, or using raw Hounsfield Units. We found that training is not sensitive to the choice of preprocessing, with all methods leading to approximately the same results, while histogram equalization being slightly worse. For the experiments described in this work, we clipped the intensities to be inside the $5$ to $99.5$ percentile range, then subtract mean and divide by standard deviation. For the Coarse step, we rescale the whole image to have $1.0$mm isotropic voxel resolution using linear interpolation, which translates into a $25$x volume reduction when starting from typical $0.2$mm isotropic voxels. Then, we crop a volume of size $64\times128\times128$ randomly from the rescaled image for each iteration.

For evaluation we selected $20$ studies from the precisely labeled dataset at random, stratified by manufacturers, as a test set for all experiments (including the one trained on weakly labeled dataset only). The rest of the data was split between training and development sets with approximately $93\%$ of data going into the training set.

\subsection{Weak label preprocessing}
\label{sec:segmentationmaskspreparation}
Manually annotated coarse axial masks are sparse and collected in the form of axial bounding boxes, which is non-standard for segmentation. To create a single continuous mask for each tooth, we apply a modification of the distance transform function --- we use the average distance from each voxel to the centerline of the individual teeth.
We perform a linear combination of voxel intensities and distances as: $$energy\_intensities = intensities + k*distances,$$ where $k$ is a parameter. We have $33$ energy masks afterward and use the $argmax$ function to label each voxel with a number from $0$ to $32$, where $0$ is the background. We use unique $k$ value for each of the manufacturers of CT scans presented in our dataset. That parameter was chosen heuristically after visual validation.
 
This preprocessing results in continuous teeth masks, but with lots of artifacts and mistakes.

\subsection{Model}
\label{ssec:localization}
We formulate the problem as a $33$-way semantic segmentation, where the background and each of the possible $32$ teeth is interpreted as a separate class.

We use a VNet \cite{vnet} fully convolutional network for both Coarse and Fine models. The Coarse model has an output width of $33$, interpreted as a softmax distribution over each voxel, assigning it either to the background or one of the $32$ teeth. The Fine model has output width of 1, interpreted as the probability of assignment of a voxel to a tooth of interest.

\subsection{Loss function}
\label{ssec:lossfunction}
Let $R$ be the ground truth segmentation with voxel value $r_i$ ($0$ or $1$ for each class), and $P$ the predicted probabilistic map for each class with voxel value $p_i$. As a loss function, we use soft negative multiclass Jaccard similarity defined as:
$$L = 1 - \frac{1}{N}\sum\limits_{i=1}^{N}{\frac{p_ir_i + \epsilon}{p_i + r_i - p_ir_i + \epsilon}},$$
where $N$ is the number of classes, which in our case is $33$ for the Coarse model and $2$ for the Fine model, and $\epsilon$ is a loss function stability coefficient that helps to avoid dividing by zero. 

\subsection{Training}
\label{ssec:training}
Training was performed in $3$ distinct phases. First, we train the Coarse model on a weakly labeled dataset with target dense masks inferred from bounding boxes according to $4.1$. We trained for $100$ epochs using Adam optimizer \cite{adam} with a learning rate of $1e-4$ decayed by a factor of 10 after 50 and 75 epochs, and a batch size of 3. For testing, we use the checkpoint with the lowest recorded validation loss.

Then, we fine-tune the same model on a precisely labeled dataset, where original dense masks were downscaled to $1$mm resolution. Training setup was the same except for increasing learning rate to $1e-3$.

As we transition to the Fine step, we use the Coarse model to prepare the training dataset. First, we obtain predicted voxel-tooth assignment in coarse $1$mm resolution. Then, we select the largest connected component for each tooth's mask. Then, we find a minimal bounding rectangle that contains the tooth volume and extend it by $3$mm in each direction to account for possible Coarse model errors (the value of $3$mm was decided by hyperparameter search from $0$mm to $5$mm). Finally, we crop the resulting rectangle from the image and precise mask. In this way, we obtain both model input and target. We apply the same procedure to obtain validation and test sets.

Finally, we train the Fine model on prepared datasets. Since we use the original resolution without rescaling or additional crops, every tooth volume has a different size. We use a batch of size 1 and a learning rate of $1e-4$.

\subsection{Metrics}
\label{ssec:metrics}
For evaluation, we use an average surface distance (ASD) score in millimeters, defined as the average of all distances from all points on the boundary of the predicted mask to the closest point on the boundary of the ground truth mask, and vice versa. \cite{asd}
To measure whole-tooth localization performance, we also measure binary voxel-wise intersection over union (IoU) between the ground truth volumetric mask and the model prediction.

\section{Results}
\label{sec:results}
As shown in table 1, for the Coarse step using full precisely labeled dataset (2) leads to significantly worse results than using full weakly labeled dataset (1), while being more expensive in terms of annotation time. This indicates that 97 CBCT scans in train set are insufficient to train good tooth segmentation model. At the same time, using only $5\%$ precisely labeled samples while fine-tuning from a model trained on full weakly labeled dataset (3.1) significantly improves results compared to both (2) and (1), while requiring only $8\%$ more annotation time then (1) and $36\%$ less annotation time then (2). Adding a Fine step model to the same setup (4.1) further improves performance by as much as $53.5\%$ in terms of ASD and by $35.8\%$ in terms of IoU without additional annotation time cost. Curiously, using only $5\%$ of precise labels with the Fine step (4.1) leads to better performance than using all precise labels without it (3.5).

Using full datasets with the Coarse-to-Fine pipeline (4.5) we can achieve $0.17$ ASD and $0.94$ IoU. At that point, performance keeps improving, indicating that the pipeline is not yet dataset-saturated.

The major problem of training on weak labels only (1) is a poor delineation between the background and nearby bones from teeth. These problems are clearly visible in roots and bitewing areas. Moreover, this model often mislabels voxels of one tooth as belonging to a neighboring tooth. The Coarse model with fine-tuning on precise labels (3) gets accurate results in such difficult cases. The Coarse model trained only on the small precise dataset (2) achieves significantly worse results than both weak and fine-tuned models, indicating insufficient dataset size. The Fine model (4) produces voxel-perfect masks, visibly very close to ideal.

\section{Conclusion}
\label{sec:conclusion}
In this work, we present two datasets of CBCTs labeled for tooth segmentation and numbering. We also present the Coarse-to-Fine segmentation pipeline with weakly supervised pretraining, achieving $0.17$mm ASD and $0.94$ IoU and showing significant improvements over several typically used baseline setups.

We show that using a coarse-to-fine framework is effective for handling large volumetric images, even when localizing and segmenting as many as $32$ small anatomical structures, at least for tooth segmentation. Our results also indicate that the strategy of collecting weak, cheap labels first is a viable approach for problems where precise labels are expensive (such as volumetric segmentation of medical images), or where proof of concept is required. These weak labels can be reused for training precise models afterwards.

\small

\bibliographystyle{IEEEbib}
\bibliography{main}

\end{document}